\documentclass{article}
\usepackage{spconf,amsmath,graphicx}

\usepackage{multirow}
\usepackage{graphicx}
\usepackage{subfigure}
\usepackage{threeparttable}
\usepackage{algorithm}
\usepackage{algpseudocode}
\usepackage{bm}
\usepackage{caption}
\usepackage{makecell}
\usepackage{algorithmicx}
\usepackage{hyperref}
\usepackage{url}

\usepackage{color}
\usepackage{enumitem}
\usepackage[skip=5pt]{caption}
\usepackage{booktabs}
\newcommand{\secref}[1]{Section \ref{#1}}
\newcommand{\figref}[1]{Figure \ref{#1}}

\usepackage{txfonts}
\let\mathbb=\varmathbb
\DeclareSymbolFont{letters}{OML}{ztmcm}{m}{it}
\title{PK-Chat: Pointer Network Guided Knowledge Driven Generative Dialogue Model}

\name{Cheng Deng\textsuperscript{1}, Bo Tong\textsuperscript{1}, Luoyi Fu\textsuperscript{1}, Jiaxin Ding\textsuperscript{1}, Dexing Cao\textsuperscript{1}, Xinbing Wang\textsuperscript{1}, Chenghu Zhou\textsuperscript{1,2}}
\address{\textsuperscript{1} Shanghai Jiao Tong University \\ \textsuperscript{2} Institute of Geographical Science and Natural Resources Research, Chinese Academy of Sciences \\ \{davendw, bool\_tbb, yiluofu\}@sjtu.edu.cn}

\begin{document}
\begin{sloppy}
\small
% \ninept
\maketitle

\begin{abstract}
In the research of end-to-end dialogue systems, using real-world knowledge to generate natural, fluent, and human-like utterances with correct answers is crucial. 
However, domain-specific conversational dialogue systems may be incoherent and introduce erroneous external information to answer questions due to the out-of-vocabulary issue or the wrong knowledge from the parameters of the neural network. 
In this work, we propose \textbf{PK-Chat}, a \textbf{P}ointer network guided \textbf{K}nowledge-driven generative dialogue model, incorporating a unified pretrained language model and a pointer network over knowledge graphs. The words generated by PK-Chat in the dialogue are derived from the prediction of word lists and the direct prediction of the external knowledge graph knowledge. Moreover, based on the PK-Chat, a dialogue system is built for academic scenarios in the case of geosciences. Finally, an academic dialogue benchmark is constructed to evaluate the quality of dialogue systems in academic scenarios and the source code is available online.~\footnote{\url{https://github.com/iiot-tbb/Dialogue\_DDE}}
\end{abstract}
\begin{keywords}
Dialogue System, Pointer Network, Academic Knowledge Graph, Natural language Generation
\end{keywords}

\section{introduction}
\label{sec:intro}
% From the initial rule-based matching dialogue understanding to intention recognition and slot extraction of user's discourse through deep learning, as well as the development of retrieval, the pipeline to generative, wide-area oriented dialogue systems continues to evolve.~\cite{Ni2022RecentAI} Due to the lack of flexibility of the pipeline approaches, generative end-to-end approaches are widely studied.
 Making a Dialogue system by using language models such as PLATO and GPT~\cite{Bao2020PLATOPD,Radford2018ImprovingLU} is becoming a mainstream research direction, finetuning on such models can generate human-like conversational responses. 
However, existing generative dialogue systems focus on providing general-purpose responses which could result in a lack of domain expertise and semantic coherence in the responses~\cite{Wang2021ImprovingDR}.  
% In addition, existing dialogue systems are mainly for the general purpose, and chatbots dedicated to domain-specific scenarios are rare.~\cite{Chen2017ASO}
% semantic coherence
%Besides, the natural one-to-many problem of generative end-to-end dialogue systems has led to a lack of clarity in the goal of generative dialogue, often resulting in ambiguous responses to the same question, since there are not enough references to world knowledge. Consequently, end-to-end method usually contribute to the casual chatbots, and there are significant challenges if one wants to accomplish domain-specific chatbots in an end-to-end manner~\cite{Mehfooz2020MedicalCF}, even with the support of pretrained language models, since the utterances can conform to the typical syntactic structure and semantic logic but lack prior knowledge if the dialogue system is generated simply by the memorized parameters of the neural network.
%现有的端到端的生成模型存在歧义回复，没有额外知识

Knowledge graphs, such as Freebase~\cite{lehmann2015dbpedia}, Yago~\cite{fabian2007yago} are introduced into the dialogue systems~\cite{Yang2020GraphDialogIG} to compensate for the absence of domain expert knowledge. 
%By using an external knowledge graph like Freebase, Yago~\cite{lehmann2015dbpedia,fabian2007yago} as a knowledge base and introducing external knowledge into the dialogue system~\cite{Yang2020GraphDialogIG}, it can compensate for the inability of the dialogue system to generate utterances that conforms to particular domains. 
In these works, knowledge graphs are embedded into vectors in the latent semantic spaces and the embedding vectors are used to generate relevant text candidate sets~\cite{Sun2019ERNIEER}. 
However, 
the semantic coherence is neglected, since
the probability of each utterance neighbor candidates is calculated independently, without considering the relation between the candidate utterances and the contextual input~\cite{Qiu2019AreTS}.   % ?
Moreover, pretrained model acquires the position information of the referenced knowledge, and if the words of the knowledge graph do not appear or are less likely to appear in the tokenizer of the pretrained model, they are unseen knowledge and out-of-vocabulary words, and thus cannot identify the specific meaning, In this condition, the knowledge memorized by the network would be misused in the response.~\cite{Bao2021PLATO2TB}. 

 %, when choosing the most probable utterance from the neighbors in the latent semantic spaces. 
%and the agent is unable to compare the semantic coherence between each candidate utterances and the input text.~\cite{Qiu2019AreTS} 
%Most existing methods embed knowledge graphs into language models in the form of vectors and generate relevant text candidate sets through the latent space of semantics~\cite{Sun2019ERNIEER}. 

%However, such methods require a large amount of parallel corpus to ensure the knowledge graph and text embedding space are the same and require many training parameters. These may lead to misaligned knowledge, ambiguous association, and error feedback.

% 就算使用了知识图谱也会存在不对齐的情况
\begin{figure*}[h]
    \centering
    \includegraphics[width=\linewidth]{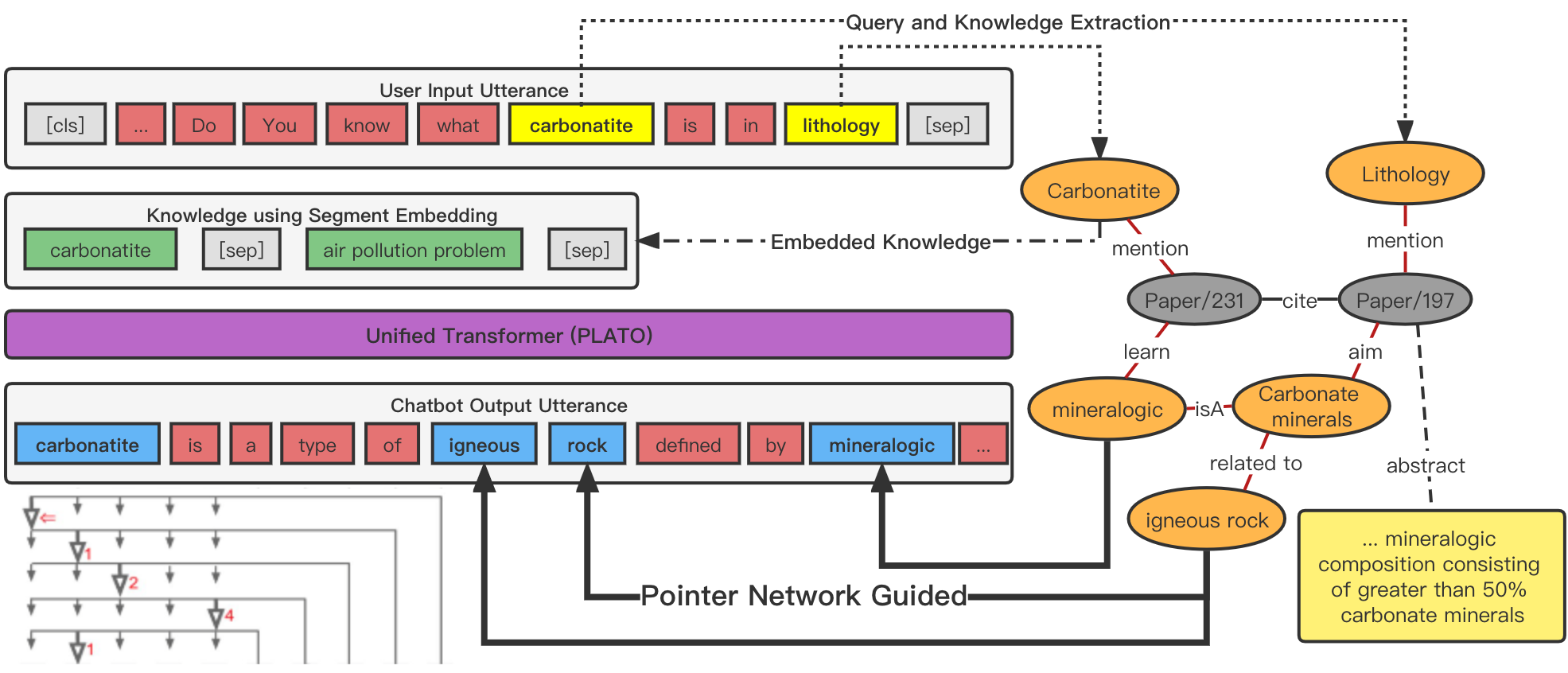}
    \caption{Overview of PK-Chat with Geoscience Academic Knowledge Graph.}
    \label{fig:overview}
    \vspace{-1.5em}
\end{figure*}

In face of the issues of knowledge driven generative dialogue model, we introduce pointer networks~\cite{Vinyals2015PointerN,See2017GetTT} to deliver information from the original input text to the output text with detailed information kept.  % to guide the knowledge to preserve from the inp.  
%To address the misaligned during the process of combine the text semantic information and the domain knowledge from a KG and avoid illogical utterance generation, using a pointer network~\cite{Vinyals2015PointerN,See2017GetTT} is an appropriate means. 
%Pointer network %, keeping as much detailed information in the original text as possible and easing the problem that the unpresented corpus is challenging to learn. 
Take text summarization as an example, where there can be words in the original corpus that has not been seen during training. For models without a pointer network, the original details are hard to restore,  
and the generated summary can  contain much inaccurate information, while with the pointer network, the details can be restored by keeping from the original text when unregistered words are encountered~\cite{Sun2018MultiSourcePN,See2017GetTT}. 

% Meanwhile, there are few datasets for particular domains, and the cost of constructing them is also high. We need to make full use of the extensibility and knowledge association of the knowledge graph, and here we propose a dialogue dataset based on the academic knowledge graph of earth sciences, which provides a new benchmark for dialogue tasks on the one hand, and facilitates us to land our model in the academic domain on the other hand.

In this work, we design a pointer network guided knowledge-driven generative dialogue model
%dialogue generation model based on a unified pretrained language model~\cite{Dong2019UnifiedLM} and pointer generation network and 
and train with the GAKG~\cite{Deng2021GAKGAM} to develop a fluent and natural knowledge-informed dialogue with users around the specific knowledge based on the corresponding geology knowledge. 
% The purpose of the system is to develop a smooth, natural, and knowledge-rich dialogue interaction with users around the specific knowledge based on the geoscience knowledge in the geoscience knowledge graph. 
The overview of the system is shown in \figref{fig:overview}. The contributions of this paper are as follows:

\begin{itemize}[leftmargin=0.8em]
    \item[1] This research paper introduces \textbf{PK-Chat}, a novel dialogue generation model that combines a pretrained language model with a pointer generation network through a flexible self-attentive mechanism. By leveraging this advanced approach, PK-Chat demonstrates superior performance compared to established baselines across various benchmarks. 
    \item[2] PK-Chat innovatively adopts the pointer network with unified pretrained language models to guide domain-specific conversation generation, a key contribution towards advancing the state-of-the-art in this domain.
    \item[3] Alongside PK-Chat, we construct GA-Dialogue, the first academic dialogue dataset with words sourced from the GAKG. The availability of this dataset represents a significant advancement, as it can be used to train other dialogue generation models, further contributing to the development of this promising field.
    % Relying on the PK-Chat and GA-Dialogue, a chatbot based on the academic knowledge graph is built and will be publicly available online.
\end{itemize}

\section{Methodology}
\label{sec:method}

In this section, we introduce \textbf{PK-Chat}, a novel dialogue generation model that leverages a unified pretrained language model and pointer generation network through a self-attentive mechanism. PK-Chat is designed to generate fluent and natural text that draws on specific domain knowledge based on academic knowledge graphs. To achieve this, PK-Chat consists of a dialogue generation model, knowledge graph retrieval, and keyword extraction subsystems that work collaboratively to produce intelligent and context-sensitive responses to user inquiries.

First, when a user inputs an utterance, the model determines whether the input is relevant to the current referenced knowledge range. Suppose the current user utterance conflict with the referenced knowledge, then the knowledge extraction algorithm will be activated to extract the text's keyword information or entity information. The graph query statement is applied to query the specific connected triples information with the keyword of the entity in the knowledge graph, and all directly connected edges and tail entities with the node will be recalled as the knowledge input part of the model, which is combined with the utterance of the user to generate a response.

% \begin{figure}[h]
%     \centering
%     \includegraphics[width=\linewidth]{ICASSP23/img/PK-CHAT.png}
%     \caption{Overview of PK-Chat model.}
%     \label{fig:PK-Chat}
%     \vspace{-1em}
% \end{figure}

\subsection{Dialogue Generation}
In order to generate reasonable dialogue responses, the generative model should fully understand the above dialogue history information and background knowledge so that the responses are accurate and consistent enough. 

Since the unification of bidirectional, unidirectional, and sequence-to-sequence objective functions enables us to straightforwardly finetune the pretrained language UniLM\cite{dong2019unified} for both NLU and NLG tasks and dialogue tasks can benefit from it, we adopted a parameter sharing self-attention mechanism transformers like UniLM-PLATO based conversational language model as the backbone of the PK-Chat, which train from social media corpus. We fine-tune it with the conversational corpus from the data in GAKG illustrated in \secref{sec:exp}. Besides, the loss functions for our task include the negative log-likelihood function used in the dialogue response generation and the pointer generation part, the bag-of-words model loss function used when predicting the words that should be in the responses, and the cross-entropy loss function used for topic switching.

\paragraph*{\textbf{Response Generation}}

For a given contextual information $c$ and a selected hidden variable $z$, the reply is given as $P(r \mid c,z)$ based on this. Where $z \in [1, K]$, each specific z value corresponds to a potential semantic behavior, and the identification of the corresponding hidden variable can be accomplished by $\operatorname{argmax}(P(z \mid r, c))$ for the given contextual information and response content.

In PK-Chat, the response generation consists mainly of discrete hidden variables, content, and knowledge information. And take the maximum likelihood estimation function as the loss function like~\autoref{gen:nll}. 
\begin{equation}
\label{gen:nll}
    \begin{aligned}
    \mathcal{L}_{N L L} &=-\mathbb{E}_{z \sim p(\mathbf{z} \mid c, k, r)} \log p(r \mid c, k, z) \\
    &=-\mathbb{E}_{z \sim p(\mathbf{z} \mid c, k, r)} \sum_{t=1}^T \log p\left(r_t \mid c, k, z, r_{<t}\right),
    \end{aligned}
\end{equation}
where $z$ is the discrete hidden variable obtained from $(c, k, r)$ and based on the probability $p(\mathbf{z} \mid c, k, r)$ are sampled. The hidden variable identification task obtains the distribution of the posterior probabilities of the hidden variables. $c$ is the conversation information above, and $k$ is the external knowledge information. And $p(\mathbf{z} \mid c, k, r)$ is a $\operatorname{softmax}$ activate function as~\autoref{gen:act},
\begin{equation}
\label{gen:act}
    p(\mathbf{z} \mid c, k, r)=\operatorname{softmax}\left(W_1 h_{[M]}+b_1\right) \in \mathbb{R}^K,
\end{equation}
where $z\in \mathbb{R}^{K}$, $h_{[M]} \in \mathbb{R}^{K}$ are the status token for the last layer of special status marker location. $W_1 \in \mathbb{R}^{K \times D}$ and $b_1 \in \mathbb{R}^{K}$ denotes the trainable parameters.

For each word $w$ in the response generation, the prediction is made by the word corresponding word list, context, and knowledge-embedded information. PK-Chat adopts the pointer network making the references to external knowledge more accurate. The probability is calculated as~\autoref{gen:prob},

\begin{equation}
\label{gen:prob}
    P(w)=\lambda_{g e n} P_{\mathrm{vocab}}(w)+\left(1-\lambda_{g e n}\right) \sum_{i: w_i=w} a_i^t,
\end{equation}

where $\lambda_{g e n}=\operatorname{sigmoid}\left(W_2 h_D+b_2\right)$, $W_2$ and $b_2$ are trainable parameters, $h_{D}$ is a hidden state of the intermediate generation result, and $a$ is denoted as the prediction of the knowledge embedding location among the context and the pointer.

In addition to the negative log-likelihood estimation of the direct task goal of generating dialogue responses, the loss function of the bag-of-words model is added to the model training process to achieve the learning of the hidden variable $z$ by predicting the words in the bag of words, specifically by predicting the words that should be in the responses through the hidden state of the last layer of $z$. Such a multi-task model can also accelerate the convergence speed of the model training. And the loss function is~\autoref{gen:lbow}.
\begin{equation}
\label{gen:lbow}
    \begin{aligned}
    \mathcal{L}_{B O W} &=-\mathbb{E}_{z \sim p(\mathbf{z} \mid c, k, r)} \sum_{t=1}^T \log p\left(r_t \mid c, k, z\right) \\
    &=-\mathbb{E}_{z \sim p(\mathbf{z} \mid c, k, r)} \sum_{t=1}^T \log \frac{e^{f_{r_t}}}{\sum_{v \in V} e^{f_v}},
    \end{aligned}
\end{equation}
where $V$ represents the size of the word list, $f$ is the softmax function $f=\operatorname{softmax}\left(W_3 h_z+b_3\right) \in \mathbb{R}^{|V|}$ that predicts the words in the target generation, and $f_{r_t}$ represents the probability value of the words generated at each moment. This prediction does not correlate to the order of each word but to the intention of making the hidden state variables capture more global information through this learning approach.

\paragraph*{\textbf{Topic Switch}}
In practice, we should select the correct external knowledge under the appropriate topic and judge whether we need to switch knowledge by comparing the current user's words and the context. Therefore, the judgment of topic switching is necessary, and we can select different knowledge at the appropriate moment. In this model, topic switching is a binary classification task to classify whether the current knowledge matches the current user utterance, and if it does, the current topic knowledge is maintained and keeps chatting on the current topic. When the current knowledge does not match the question asked by the user, the keyword extraction module is triggered, and the corresponding entity and edge information of the extracted keyword is queried in the knowledge graph. We choose the cross-entropy loss function as~\autoref{gen:lts}.
\begin{equation}
\label{gen:lts}
    \mathcal{L}_{T S}=-\log p\left(l_{\text {true }}=1 \mid k, c, r^{+}\right)-\log p\left(l_{\text {true }}=0 \mid k, c, r^{-}\right).
\end{equation}
Give the knowledge during the dialogue with $l_{\text {true }} = 1$, randomly sample the knowledge in the other topic, and label it as $l_{\text {true }} = 0$. Overall, the loss function of the whole model is:
\begin{equation}
    \mathcal{L}=\mathcal{L}_{N L L}+\mathcal{L}_{B O W}+\mathcal{L}_{T S},
\end{equation}
where $\mathcal{L}_{N L L}$ acts directly on the generation purpose, $\mathcal{L}_{B O W}$ acts on the hidden state learning and assists in the generation task. The $\mathcal{L}_{T J}$  is used for topic classification, so the whole model uses a multi-task learning method.

\subsection{Keyword Extraction}
When the user's utterances mention entities that are in the knowledge graph, the critical information will be extracted via rule-based keyword extraction method, TF-IDF~\cite{Ramos2003UsingTT}, TextRank~\cite{Mihalcea2004TextRankBO} and BiLSTM+CRF~\cite{Huang2015BidirectionalLM} NER methods to extract the current entity during the communication with the user.

\begin{itemize}[leftmargin=0.8em]
    \item We use a rule-based method by constructing regular expressions like \textit{``(what$\mid$which$\mid$where)(is$\mid$are)(the)[a-z]\{0,5\}?''} to match the questioning phrase, which can quickly locate the corresponding keyword.
    \item We use TF-IDF and TextRank to obtain the most important words by multiplying the word frequency of a word and its inverse document frequency to indicate the importance of a word.
    \item We also use the BiLSTM-CRF model that defines the knowledge information extraction of user conversations as a sequence annotation task for keyword extraction.
\end{itemize}

When it comes to the keyword extraction training data, TF-IDF and TextRank are unsupervised methods that do not require the construction of labeled data for training, so the dataset construction for information extraction is mainly to enable the BILSTM+CRF model to have a good performance on this keyword extraction task. According to the characteristics of the dialogue data in this paper, the entity information in the dialogue is a reference to the entities in the knowledge graph, so the annotation task in this part does not need a large amount of manual annotation, and we only need to search and locate the entities in the dialogue and do the automatic annotation.

\begin{table*}[h]
    \resizebox{\textwidth}{!}{%
    \begin{tabular}{@{}cccccccccc@{}}
    \toprule
     &  & \multicolumn{3}{c}{Automatic Evaluation} & \multicolumn{5}{c}{Human Evaluation} \\ \midrule
    Dataset & Model & BLEU-1/2 & Distinct-1/2 & Knowledge R/P/F1 & Readability & Relevance & Consistency & Informativeness & Naturalness \\ \midrule
    \multirow{3}{*}{GA-Dialogue (part1)} & PLATO (Unidirect) & 0.054/0.042 & 0.099/0.270 & 0.002/0.011/0.003 & 0.60 & 0.50 & 0.502 & 0.40 & 0.37 \\
     & PLATO & 0.415/0.354 &\textbf{ 0.165}/0.361 & 0.099/0.218/0.124 & 2.67 & 2.10 & 2.23 & 2.37 & \textbf{2.20} \\
     & PK-Chat (Ours) & \textbf{0.636/0.532} & 0.139/\textbf{0.366} & \textbf{0.100/0.228/0.128} & \textbf{2.73} & \textbf{2.26} & \textbf{2.40} & \textbf{2.43} & 1.90 \\ \midrule
    \multirow{3}{*}{GA-Dialogue (part2)} & PLATO (Unidirect) & 0.106/0.086 & 0.050/0.137 & 0.002/0.048/0.003 & 0.00 & 0.03 & 0.03 & 0.03 & 0.03 \\
     & PLATO & 0.342/0.268 & \textbf{0.105/0.322} & 0.022/0.246/0.040 & \textbf{2.93} & \textbf{2.53} & 2.50 & 2.63 & 2.33 \\
     & PK-Chat (Ours) & \textbf{0.496/0.383} & 0.065/0.237 & \textbf{0.044/0.273/0.074} & 2.80 & 2.43 & \textbf{2.57} & \textbf{2.80} & \textbf{2.60} \\ \midrule
    \multirow{3}{*}{Persona-Chat} & PLATO (Unidirect) & - & 0.003/0.010 & 0.018/0.084/0.028 & 0.30 & 0.24 & 0.20 & 0.27 & 0.13 \\
     & PLATO & 0.231/0.178 & 0.014/0.053 & \textbf{0.028/0.138/0.044} & \textbf{2.83} & \textbf{2.17} & \textbf{2.30} & 1.97 & 2.33 \\
     & PK-Chat (Ours) & \textbf{0.257/0.199} & \textbf{0.015/0.062} & 0.026/0.131/0.042 & 2.57 & 1.96 & 2.17 & \textbf{2.03} & \textbf{2.34} \\ \midrule
    \multirow{3}{*}{DailyDialog} & PLATO (Unidirect) & 0153/0.117 & 0.042/0.153 & - & 2.71 & 1.97 & 1.67 & 1.33 & 1.27 \\
     & PLATO & 0.388/0.304 & \textbf{0.055/0.303} & - & 2.57 & 2.33 & 2.10 & 2.07 & 1.90 \\
     & PK-Chat (Ours) & \textbf{0.416/0.329} & 0.049/0.282 & - & \textbf{2.77} & \textbf{2.71} & \textbf{2.53} & \textbf{2.63} & \textbf{2.41} \\ \bottomrule
    \end{tabular}%
    }
    \caption{Comparison with baselines.}
    \label{tab:exp}
    \vspace{-1em}
\end{table*}

\subsection{Retrieve over Knowledge Graph}
In order to ensure the efficiency of the knowledge retrieval, we choose a reasonable storage method for the external knowledge graph. In this paper, we choose GAKG, an academic knowledge graph in geoscience, so as to deploy an academic dialogue system.

The GAKG is a collection of papers' illustrations, text, and bibliometric data, is currently the largest and most comprehensive geoscience academic knowledge graph, consisting of more than 120 million triples with 11 kinds of concepts connected by 19 relations, stored in RDF format. We download the full copy of GAKG and store it in the graph database (Neo4J). After that, we build a GA-Dialogue dataset to train an academic chatbot. First, we randomly sampled all the information about the connected edges and tail entities of a single head entity on the knowledge graph of GAKG, constructed a specific dialogue scenario based on the sampled information, and started a specific dialogue around the information of the entity, i.e., we quoted the information of the entity in the dialogue to reply. In order to improve the quality of the dialogue dataset, we invited 20 geographers who understand the detail of GAKG to participate in the construction and let them retain the label format of the entity. Five hundred fifty dialogue scenarios and 3,615 dialogues were constructed in GA-Dialogue. The average number of utterances of users per scenario is \textbf{6.7}.

However, the number of dialogue datasets is not large enough, so we increase the number of data by cleaning and constructing the public dataset. We used the Baidu DuConv~\cite{Wu2019ProactiveHC} and Baidu DuRecDial~\cite{Liu2020TowardsCR} dialogue datasets as external datasets to introduce. For the DuConv dataset, there are 29,858 conversations in the scenes, with an average of \textbf{9} rounds of conversation per scene.
In order to unify the data in this dialogue dataset with the dialogue data in our GAKG. There are two types of knowledge in the conversation dataset: conversation goals, and knowledge. We integrate the conversation goals and knowledge aggregated in the conversation dataset into the knowledge as the unified external knowledge. For the DuRecDial dataset, there are 10,200 conversations in the scenes, with an average of \textbf{15} rounds of conversation per scene. There are three types of knowledge useful in the conversation dataset: conversation goals, knowledge, and user profile. we integrate the conversation goals, knowledge, and user profile aggregated in the conversation dataset as the knowledge. In this way, the data format is aligned with the conversation format of GAKG.

We sampled a few data and finally, GA-Dialogue has 1,000 dialogue scenarios and 8219 dialogue rounds, with an average of \textbf{8.21} dialogue rounds per scenario.

\section{Experiment}
\label{sec:exp}
In this section, we evaluate the automatic evaluation results and the human evaluation results of the model of the dialogue system. This section details the models' benchmarks and evaluation metrics in experimental setup and evaluation results. 
\vspace{-1em}
\subsection{Experimental Setup}
In this subsection, we briefly introduce the benchmarks, baselines, and metrics we selected to do experiments on our model.

\paragraph*{Benchmarks}
We choose Persona-Chat~\cite{Zhang2018PersonalizingDA} and DailyDialog~\cite{Li2017DailyDialogAM} as the general benchmark, and we build an Academic Knowledge-Graph-based dialogue benchmark GA-Dialogue. 
\begin{itemize}[leftmargin=0.8em]
    \item Persona-Chat is a dataset of knowledge-based conversations on persona profiles (background knowledge).
    \item DailyDialog is a chitchat dataset containing high-quality human conversations about everyday life.
    \item GA-Dialogue (Part 1 \& 2), we divide the GA-Dialogue test dataset into two parts, the first part (part1) of the set contains dialogue data on the knowledge graph of GAKG, and the second part (part2) of the dialogue evaluation comes from the evaluation of the dialogue data introduced by external dataset. The datasets are available at Github Repo.\footnote{\url{https://github.com/davendw49/PK-Chat}}
\end{itemize}

\paragraph*{Baselines}
We choose the PLATO~\cite{Bao2020PLATOPD} and PLATO (Unidirect) as the baselines since PLATO achieves the sota results for the known model size of the same scale and PLATO (Unidirectional) is chosen as the baseline model to analyze the effect of the unidirectional attention mechanism on the final generation of the model, it is consistent with the GPT~\cite{Radford2019LanguageMA} series in the model self-attention structure. 

\paragraph*{Metrics}
Different from task-oriented dialogue systems, open-domain dialogue systems are complicated to evaluate the performance of dialogue systems through a specific metric due to the flexibility of the dialogue. In general, the open-domain dialogue systems are measured through objective and subjective evaluations, and the automatic and human evaluation methods used in this paper compare each model.

We choose BLEU~\cite{Chen2014ASC} (bilingual evaluation understudy), Distinct~\cite{Li2016ADO} and Knowledge~\cite{Bao2020PLATOPD} as the \textbf{Automatic Evaluation Metric}, and greater the metrics are the better the models perform.
\begin{itemize}[leftmargin=0.8em]
    \item BLEU is used for the evaluation of the generation task determined by calculating the overlap between the generated responses and the $n$-gram of the tags. In this paper, we set $n$ as $1$ and $2$. 
    \item Distinct is set up for the measurement of diversity rubric for evaluating generated sentences by counting the ratio of unique $n$-gram of the words. In this paper, we set $n$ as $1$ and $2$.
    \item Knowledge is used to determine whether the cited knowledge is correct or incorrect.
\end{itemize}

For the human evaluation method, the human evaluation includes five indicators as described in~\cite{Liang2021TowardsSC}, and we use them as the \textbf{Human Evaluation Metrics} in this paper: Readability, Relevance, Consistency, Informativeness, and Naturalness. In each benchmark, 500 generated dialogues and their contextual information are randomly selected as evaluation data, and 20 geoscientists are invited to analyze the dialogue performance evaluation and score them from [0,1,2,3] points in each of the above five aspects.

\subsection{Experimental Result}

In GA-Dialogue (part1), the PK-Chat model outperforms the PLATO model in BLEU, Distinct, and Knowledge metrics, and in GA-Dialogue (part2), it outperforms the PLATO model and the baseline model in BLEU metrics and Knowledge metrics. The PK-Chat outperforms the PLATO model in all five dimensions of the human evaluation metrics, with four of the highest metrics in the first part of the evaluation set and three of the highest in the second part of the evaluation set. Thus, both the automatic and human evaluation metrics have improved.

Similarly, in the Persona-Chat dataset, the PK-Chat model outperforms the baseline models on the automatic measures BLEU and Distinct but slightly underperforms the PLATO model on the Knowledge measure. The PK-Chat model outperforms the PLATO model in terms of the human evaluation metrics, and our proposed method does not have any gain on the final dialogue generation in this part of the dialogue dataset, since the dataset references the knowledge of the user task portrait, of the knowledge part is rarely directly referenced when answering the questions.

As for DailyDialog, the proposed method in this paper outperforms the baseline model on this dataset for automatic and human evaluation metrics. For the information whose context is a historical conversation, the model in this paper can enhance the metrics. By observation on the dataset, compared with the Persona-Chat, the conversation content usually revolves around the same topic, and the coherence between the conversation and context of the chitchat is stronger, so the model can replicate the learning of the words in the previous question through the pointer network so that the model will have a good performance effect.

\section{Conclusion}
This paper proposes PK-Chat, a knowledge graph-enhanced model via a unified pretrained language model and pointer generation network to realize academic dialogues, aiming to develop a fluent, natural, and knowledge-informative dialogical interaction with scholars. By combining a unified pretrained language model and a pointer network, the model could accurately refer to the knowledge mentioned in the KGs. Moreover, we put forward a GA-Dialogue as a benchmark to evaluate dialogue agents. 
% Lastly, we deploy the model over a geoscience academic knowledge graph and provide a user-friendly chatbot online.

\bibliographystyle{IEEEbib}
\bibliography{refs}
\end{sloppy}
\end{document}